\documentclass{article}
\usepackage{ijcai25}

\usepackage{times}
\usepackage{soul}
\usepackage{url}
\usepackage[hidelinks]{hyperref}
\usepackage[utf8]{inputenc}
\usepackage[small]{caption}
\usepackage{graphicx}
\usepackage{amsmath}
\usepackage{amsthm}
\usepackage{booktabs}
\usepackage{algorithm}
\usepackage{algorithmic}
\usepackage[switch]{lineno}
\usepackage{multirow, colortbl}
\usepackage{adjustbox}
\usepackage{amssymb}
\usepackage{colortbl}
\usepackage{enumitem}
\usepackage[svgnames]{xcolor}
\title{CrossVTON: Mimicking the Logic Reasoning on Cross-category Virtual Try-on guided by Tri-zone Priors}
\author{
Donghao Luo$^{1,2*}$\and
Yujie Liang$^{3*}$\and
Xu Peng$^2$\and
Xiaobin Hu$^2$\and
Boyuan Jiang$^2$\and
Chengming Xu$^2$\and
Taisong Jin$^3$\and
Chengjie Wang$^2$\and
Yanwei Fu$^{1\dagger}$\\
\affiliations
$^1$Fudan University\and
$^2$Tencent\and
$^3$Xiamen University\\
\emails
dhluo24@m.fudan.edu.cn,
liangyujie@stu.xmu.edu.cn,
\{ppxupeng, xiaobinhu, byronjiang, chengmingxu, jasoncjwang\}@tencent.com,
jintaisong@xmu.edu.cn,
yanweifu@fudan.edu.cn
}
\begin{document}

% \maketitle

\twocolumn[{%
\renewcommand\twocolumn[1][]{#1}%
\maketitle
% \vspace{-35pt}
\begin{center}
    \centering
    \captionsetup{type=figure}
\includegraphics[width=0.9\textwidth]{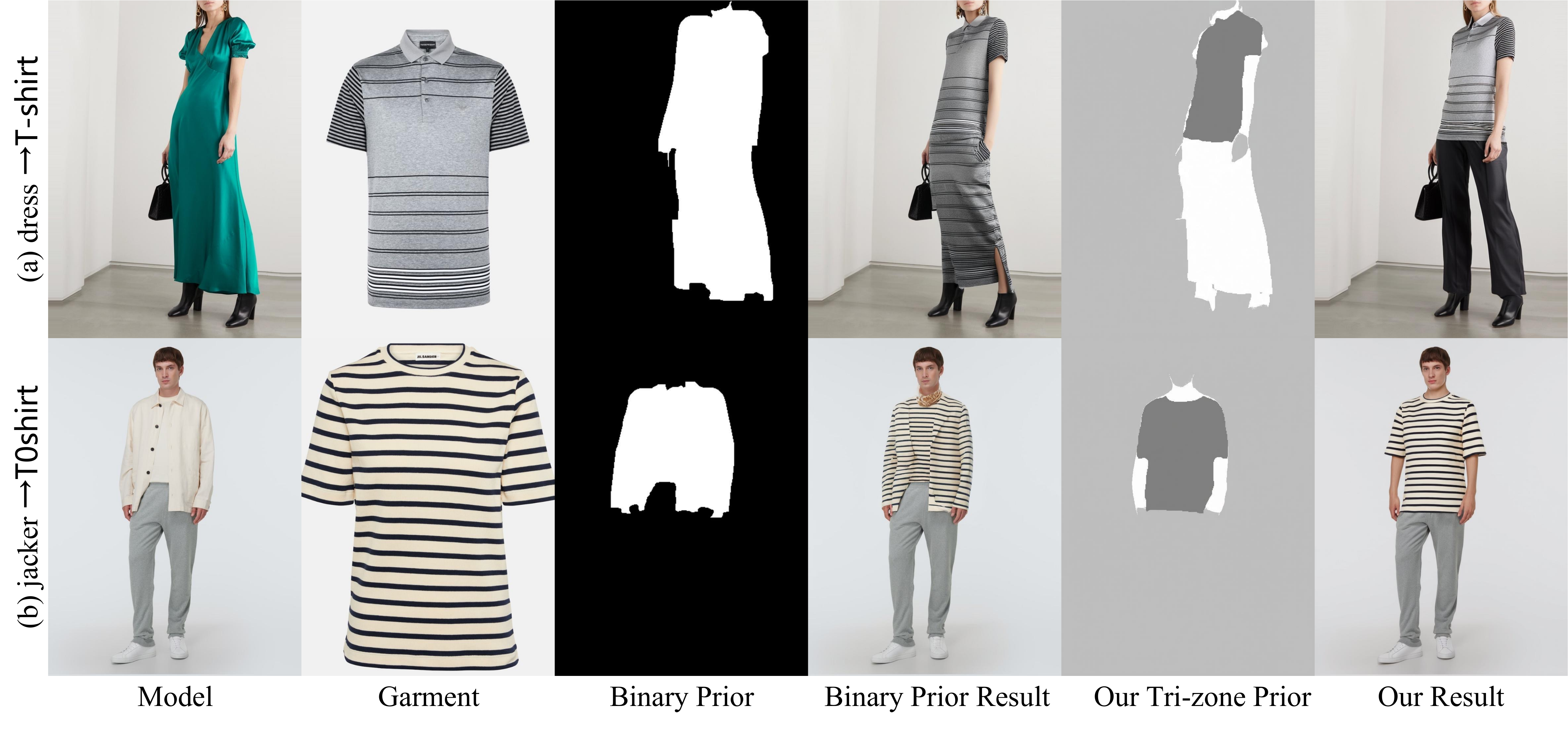}
    \vspace{-8pt}
    % \captionof{figure}{\small  Green-screen objects with matting-level annotations generation by our DiffuMatting, including nets, grid and semitransparent tough objects and extended to almost any class (\textit{e.g.,} Transportation, Architecture, Toy, \textit{etc}) without any parameters fine-tuning.
    % }
    \vspace{-2mm}
    \captionof{figure}{\small Tri-zone prior decomposes the content of the model image to determine whether it belongs to the try-on, reconstruction, or imagination zone. Different from commonly used binary mask priors, such priors can endow our CrossVTON with the capability of cross-category virtual try-on.
    % FiTDiT shows superior virtual try-on for texture-aware maintenance and  size-aware fitting challenge across any scenario.
    }
    % \caption{Figure 1. Sample images from our DIS5K dataset.}
    \label{fig:teaser}
    \vspace{-2mm}
\end{center}%
}]

\if TT\insert\footins{\noindent\footnotesize{
$*$ equal contribution, ${\dagger}$ corresponding author\\
% Project page: \url{}.
}}\fi

\begin{abstract}

Despite remarkable progress in image-based virtual try-on systems, generating realistic and robust fitting images for cross-category virtual try-on remains a challenging task. The primary difficulty arises from the absence of human-like reasoning, which involves addressing size mismatches between garments and models while recognizing and leveraging the distinct functionalities of various regions within the model images.
To address this issue, we draw inspiration from human cognitive processes and disentangle the complex reasoning required for cross-category try-on into a structured framework. This framework systematically decomposes the model image into three distinct regions: try-on, reconstruction, and imagination zones. Each zone plays a specific role in accommodating the garment and facilitating realistic synthesis.
To endow the model with robust reasoning capabilities for cross-category scenarios, we propose an iterative data constructor. This constructor encompasses diverse scenarios, including intra-category try-on, any-to-dress transformations (replacing any garment category with a dress), and dress-to-any transformations (replacing a dress with another garment category). Utilizing the generated dataset, we introduce a tri-zone priors generator that intelligently predicts the try-on, reconstruction, and imagination zones by analyzing how the input garment is expected to align with the model image.
Guided by these tri-zone priors, our proposed method, CrossVTON, achieves state-of-the-art performance, surpassing existing baselines in both qualitative and quantitative evaluations. Notably, it demonstrates superior capability in handling cross-category virtual try-on, meeting the complex demands of real-world applications.

% FitDiT surpasses all baselines in both qualitative and quantitative evaluations. It excels in producing well-fitting garments with photorealistic and intricate details, while also achieving competitive inference times of 4.57 seconds for a single $1024\times 768$ image after DiT structure slimming, outperforming existing methods. The code and dataset will be made publicly available.

    %% considering this tri-zone priors, we 
    % To embed such a logical chain that explicitly 
    
    % explicitly visualize three areas, we propose a 
    % tri-zone priors generator to conform to
    %%%数据构造：to build up an iterative data constructor in keeping with the human-mimic brain on cross-categolory try-on. 
    %%短换长的时候可以，长换短的时候需要脑补合适的区域
\end{abstract}

\section{Introduction}
The significant expansion of e-commerce has led to a continuous demand for a more convenient and customized shopping experience. Image-based virtual try-on (VTON) has become a popular method for producing realistic images of human models dressed in particular clothing items, thus improving the consumer shopping experience. Recently, numerous researchers ~\cite{dong2019towards,ge2021disentangled,issenhuth2020not,han2019clothflow,han2018viton,he2022style,minar2020cp,wang2018toward,yang2020towards}
have devoted considerable effort to attaining more lifelike and photorealistic virtual try-on outcomes. 

As a predominant generative method, Diffusion \cite{rombach2022high} shows high authenticity generation and rich texture-wise maintenance compared with generative adversarial networks (GANs) \cite{goodfellow2020generative}. Despite these advancements, Diffusion-based approaches often struggle with the cross-category virtual try-on \cite{kim2024stableviton,sun2024outfitanyone,morelli2023ladi,gou2023taming} (\textit{e.g.,} long skirt $\leftrightarrow$ upper jacket, short sleeves $\leftrightarrow$ long skirt) widely existing in the real-world scenarios. Although Anyfit \cite{li2024anyfit} and AVTON \cite{liu2024arbitrary}
% and FitDit \cite{jiang2024fitdit} 
have attempted to mitigate the cross-category virtual try-on, they mainly adopt the adjustment on the mask strategy to fit the length of target garments. To handle the challenging cross-category cases (\textit{e.g.,} long skirt $\leftrightarrow$ upper jacket), these methods usually require users to specify one of three major categories (\textit{i.e.,} upper garments, lower garments, and dresses) to distinguish the area to be processed, rather than adaptively determine the area based on the model image and garment image. Such manual intervention still suffers from the performance deterioration and the lack of reasoning ability to reasonably paint the imagination zone, as shown in Fig. \ref{fig:teaser}.
% and fail  requiring the model to  paint the imagination zone. 
% These methods usually require users to specify one of three major categories, namely upper garments, lower garments, and dresses, to distinguish the area to be replaced, rather than adaptively determine the area based on the model image and garment image. 
% For the situation of replacing an upper garment or trousers with a dress, it can be generated by specifying the dress category and  replacing full-body area. However, when a user uploads a model wearing a dress and wants to replace it with a T-shirt, harmonious replacement cannot be achieved even by specifying the category. Even when trying on clothes from other subcategories within the same major category, there will still be problems. For example, when a model is wearing a short T-shirt and wants to replace it with a mid-long trench coat or vice versa, due to the mismatch of the mask, the result generated for the former usually has a shorter trench, while the latter usually stretches the T-shirt to fill the trench area.

The hidden reason for this obstacle lies in the lack of logical reasoning about how the target garment is dressed on the model and split the picture into the different zones, \textit{i.e.,} the reconstruction zone that is learning from the model image, the try-on zone that is referring to the garment image, and imagination zone that developing the imaginative potential of Diffusion to paint on. Thus, to enhance cross-category try-on performance in more practical and diverse real-world scenarios, it is both meaningful and essential for the try-on model to develop logical reasoning and adaptively learn the variable tri-regions based on the input garment and model images.

Following this research line, we propose a novel pipeline to prompt the tri-zone priors via mimicking logic reasoning for cross-category try-on. Such a pipeline includes the iterative cross-category data construction and progressively training manner guided by tri-zone priors.  
Basically, we classify the cross-category cases into the intra-category, Any-to-Dress and Dress-to-Any cross-category depending on the length and category of the target garments. Given a model and a garment, we customize a tri-zone priors generator to reason the different function zones as a constraint for better reasonable cross-category try-on. Typically, the imagination region is required when the clothing in the original model image does not fully cover the target garment area. By leveraging the masks of these three regions, it is possible to eliminate the need for manual specification of the clothing category in the virtual try-on task, thereby facilitating cross-category try-on. 
% Additionally, these three types of regions can be adaptively determined based on the model image and the target clothing. 

Considering that the existing models fail to directly construct challenging dress-to-any cross-category cases  (\textit{e.g.,} long skirt $\to$ upper or lower), we tailor an iterative cross-category data construction to progressively compose the quadruplets data from simple to more complex cases. 
The iterative cross-category data construction is first based on the off-the-shelf try-on model and then adopts the mask-adjust operator (\textit{i.e.,} {stretch or shorten mask}) to collect intra-category data (\textit{e.g.,} upper $\to$ upper) and Any-to-Dress cross-category data (\textit{i.e.,} change the any category with dress) to compose quadruplets data (\textit{e.g.,} synthetic model with long-skirt, short garment, real-model with short garment, tri-zone prior ground-truth).  
After train the CrossVTON with the ability with the intra-category and Dress-to-Any cross-category try-on, we then use this pre-trained CrossVTON to obtain the Dress-to-Any cross-category quadruplets data  (\textit{i.e.,} real model with long-skirt, long garment, synthetic-model with short garment, tri-zone prior ground-truth). 
With the aid of such an iterative cross-category data construction, we finally progressively train two-stage CrossVTON to both satisfy the intra-category and cross-category virtual try-on. 
To mitigate the performance deterioration caused by the synthesis image, we always obey the principle that the real image is regarded as the ground-truth and the synthesis image is only as the input. 
Such a principle encourages the virtual try-on results close to the real data distribution. 
Overall, the main contributions of this paper can be summarized as follows:
\vspace{-1mm}
\noindent
\begin{itemize}[leftmargin=*]
\item A novel tri-zone priors are proposed to mimic the logic reasoning to distinguish different functionalities of various zones (\textit{i.e.,} try-on, reconstruction, or imagination area) after considering the cross-category inputs.
\vspace{-1mm}
\item An iterative cross-category data scheme is designed to successively generate quadruplets data for cross-category virtual try-on.
\vspace{-1mm}
\item A progressively training manner guided by tri-zone priors to enable the CrossVTON with the capability of cross-category virtual try-on.
\vspace{-1mm}
\item Extensive qualitative and quantitative evaluations have clearly demonstrated CrossVTON's superiority over state-of-the-art virtual try-on models, particularly in managing cross-category virtual try-on scenarios.
\end{itemize}

\section{Related Work}

\noindent\textbf{Image-based Virtual Try-on}.
GAN-based methods \cite{lee2022high,men2020controllable,xie2023gp,yang2023occlumix} have been extensively explored for natural image generation but often struggle to produce high-fidelity outfitted images. With the rapid progress of Text-to-Image diffusion models \cite{saharia2022photorealistic,ruiz2023dreambooth,hu2024diffumatting}, recent studies \cite{chen2024wear,liang2024vton,kolors,zhu2023tryondiffusion} have adopted pre-trained diffusion models as generative priors for virtual try-on.
Approaches like IDM-VTON \cite{choi2024improving} and FitDiT \cite{jiang2024fitdit} refine inpainting diffusion models to preserve details and enhance image realism. However, despite their success in generating high-quality results, these methods face significant challenges in cross-category virtual try-on due to the lack of priors necessary for reasoning about size mismatches, a key challenge in real-world applications. Motivated by this, we present a novel CrossVTON method here.

\noindent\textbf{Cross-category Virtual Try-on}.
Cross-category virtual try-on remains a challenging yet under-explored problem, limiting the development of generalized solutions. AVTON~\cite{liu2024arbitrary} introduces a limbs prediction module for basic cross-category cases, while AnyFit~\cite{li2024anyfit} uses an adaptive mask boost to refine masks during inference. However, these methods only handle simple scenarios (\textit{e.g.,} long sleeves $\leftrightarrow$ short sleeves) and struggle with complex cases such as long skirts $\leftrightarrow$ upper jackets. To address this, we propose a tri-zone prior framework to emulate logical reasoning, enabling differentiation of functional zones in cross-category try-on.

% To prevent the entire inpainting area from being filled due to a strict mask strategy, FitDiT \cite{jiang2024fitdit} proposes a dilated-relaxed mask strategy to reduce garment shape leakage and enable the model to adaptively learn the overall shape of garments. 

% \noindent\textbf{Image Synthesis of Try-on}.
% Basically, the try-on training data scheme requires paired data of the same individual wearing different outfits with the same ID, pose, and background, but only the garment varying. 
% However, such triplet or even quadruplets data is difficult to collect and thus, some research have shifted their attention on how to build the well-curated data constructor to satisfy the specific try-on tasks. 
% WUTON \cite{issenhuth2020not} and PF-AFN \cite{ge2021parser} introduce a student-teacher paradigm where the teacher model is trained as a parsing-based reconstruction to guide the student model to synthesize try-on results without relying on the parsing model. 
% These studies utilize realistic try-on images as training data, but they are hindered by limitations in data scale and diversity across various scenarios.
% Recently, BooW-VTON \cite{zhang2024boow} generates pseudo training triplet pairs via the off-the-shelf IDM model and then introduce the wild data augmentation for better adopting for the mask-free virtual try-on in the wild. However, the aforementioned methods fails to handle the cross-category and size-mismatching virtual try-on data constructor. 

\noindent\textbf{Image Synthesis of Try-on}.
Try-on training typically requires paired data where the same individual appears in different outfits with identical ID, pose, and background, varying only the garment. However, such data is challenging to collect, prompting research into curated data constructors for specific try-on tasks. WUTON~\cite{issenhuth2020not} and PF-AFN~\cite{ge2021parser} adopt a student-teacher paradigm, using parsing-based reconstruction to train models without relying on parsing during inference. While these methods leverage realistic try-on images, they are limited by data scale and scenario diversity.
Recently, BooW-VTON \cite{zhang2024boow} generates pseudo triplet pairs using off-the-shelf IDM models and employs wild data augmentation for mask-free virtual try-on. However, these methods still struggle with cross-category and size-mismatching scenarios in virtual try-on data construction, which are tackled in this paper.

\begin{figure*}[htb]
    \centering
    \includegraphics[width=0.95\textwidth]{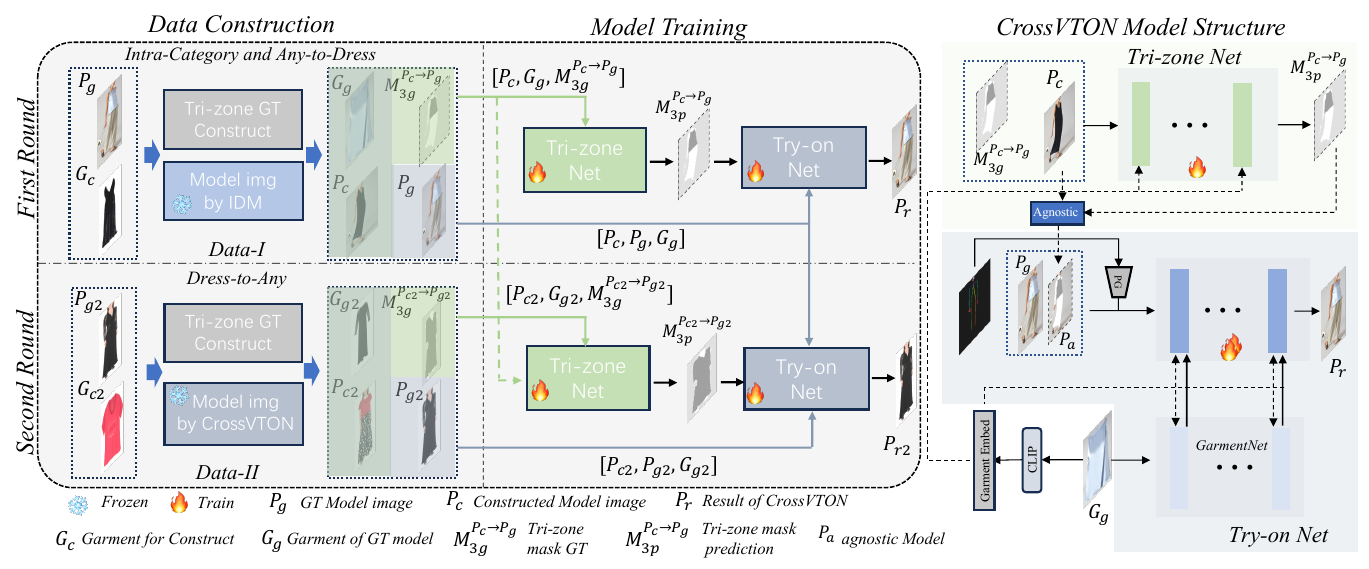}
    % \vspace{-1mm}
    \caption{An overview of the whole pipeline and the structure of CrossVTON which consists of Tri-zone  and Try-on Net. The pipeline illustrates two rounds iterative cross-category data construction by synthesizing the Intra-category, Any-to-dress, and  Dress-to-any data. At each round, the CrossVTON is trained progressively to generate tri-zone priors and endow the ability of cross-category virtual try-on.}
    \label{fig:main_framework}
    % \vspace{-1mm}
\end{figure*}

\section{Method}

% \LaTeX{} and Word style files that implement these instructions
% can be retrieved electronically. (See Section~\ref{stylefiles} for
% instructions on how to obtain these files.)

To solve cross-category try-on, we propose CrossVTON to support cross-category virtual try-on. In Section 3.1, we propose that solving this problem requires prior knowledge of three types of regions. In Section 3.2, we present CrossVTON model consisting of Tri-zone Net and Try-on Net to respectively generate and utilize tri-zone for try-on task. In Section 3.3, we introduce our progressive learning paradigm for cross-category try-on task.

\subsection{Tri-zone Priors for Virtual Try-on}
Existing diffusion-based methods typically regard virtual try-on as a conditional inpainting task. 
% They input a model image and the garment area of  the model that needs to be replaced is occluded, the garment to be changed is inputed as a reference condition to complete the occluded areas. 
The mask is only derived from the model image rather than comprehensively taking into account how the garment is dressed on the model. 
Moreover, the mask only focus on two types of zone: reconstruction region and generation region without the capability to reason the imagination zone.

% These methods usually require users to specify one of three major categories, namely upper garments, lower garments, and dresses, to distinguish the area to be replaced, rather than adaptively determine the area based on the model image and garment image. 
% For the situation of replacing an upper garment or trousers with a dress, it can be generated by specifying the dress category and  replacing full-body area. However, when a user uploads a model wearing a dress and wants to replace it with a T-shirt, harmonious replacement cannot be achieved even by specifying the category. Even when trying on clothes from other subcategories within the same major category, there will still be problems. For example, when a model is wearing a short T-shirt and wants to replace it with a mid-long trench coat or vice versa, due to the mismatch of the mask, the result generated for the former usually has a shorter trench, while the latter usually stretches the T-shirt to fill the trench area.

To tackle the aforementioned problems, we propose that virtual try-on task actually needs to distinguish three types of regions: The \textit{try-on region} $Z^{tryon}$, which indicates the area covered by the target clothing worn on the model. This area should comprehensively take into account the information of both the model and the garment. It should maintain consistency with the model in aspects of posture and body shape, and with the clothing in terms of pattern. The \textit{reconstruction region} $Z^{recon}$ which represents the area that ought to be exactly the same as the original image, typically including the face, hands, feet, and background.  The \textit{imagination region} $Z^{imagi}$ which reveals the area that needs to be complemented by the model through its imaginative faculty, and the outcome of this supplementation should be as harmonious as feasible with the other two regions.
% It is worth noting that among the three types of regions, the try-on region and the reconstruction region are essential. In most cases, the imagination region is necessary (when the clothing in the original model picture fails to fully cover the new clothing area), while in certain instances, it can be dispensed with (when the clothing in the original model picture can entirely cover the new clothing area). By leveraging the mask of these three regions, it becomes feasible to avoid manual specification of the clothing category in the virtual try-on task and accomplish cross-category try-on. Moreover, these three types of regions must be adaptively determined through the model image and the target clothing.
It is worth noting that among the three types of regions, the try-on region and the reconstruction region are essential. Typically, the imagination region is necessary when the clothing in the original model image does not fully cover the new clothing area. 
% However, in certain instances, it can be omitted if the clothing in the original model image completely covers the new clothing area. 
By utilizing the masks of these three regions, it becomes possible to avoid manually specifying the clothing category in the virtual try-on task, thereby enabling cross-category try-on. Furthermore, these three types of regions should adaptively determined based on the model image and the target clothing.

\subsection{Model Structure of CrossVTON}
As shown in Fig. \ref{fig:main_framework}, we design a two-stage pipeline, termed as CrossVTON for cross-category  virtual try-on. For the first stage, a tri-zone mask is generated via Tri-zone Net to distinguish three types of regions. Subsequently, in the second stage, leveraging the tri-zone mask as a prior, the process of try-on is executed in a manner of inpainting via Try-on Net.

The objective of the Tri-zone Net is to generate a reasonable tri-zone mask given a model and target clothing. 
As shown in Fig. \ref{fig:main_framework}, we adopt SD3 as the backbone. To more effectively extract high-level semantic information such as the pattern and length of the clothing, we utilize the image encoder of CLIP to extract clothing features and then substitute the text embedding in the original SD3 with these features. 
The goal of the Try-on Net is to obtain a natural and harmonious try-on result given a model, target clothing, and the tri-zone prior. 
As depicted in Fig. \ref{fig:main_framework}, we also adopt SD3 as backbone. 
Since the focus of the second stage is on maintaining the details of the clothing, which corresponds to low-level features, we employ GarmentNet to extract clothing features. The structure of GarmentNet is identical to that of SD3, and it is initialized with the weights of SD3. Subsequently, the clothing feature is fused to latent feature by concatenating  K and V of the clothing features and then calculating self-attention. The training loss for first and second stage are both consistent with that of SD3,
\begin{equation}
\begin{aligned}
\label{eq:garment_loss}
    L_{g}&=\mathbb{E}_{\epsilon \sim \mathcal{N}(0,1), t \sim \mathcal{U}(t)}[w(t)||\epsilon_\theta(z_{t};I_{vec},t)-\epsilon||^2],
\end{aligned}
\end{equation}
where $z_{t}=(1-t)z_0+t\epsilon$ is the noisy latent, and $w(t)$ is a weighting function at each timestep $t$. 

% \textcolor{red}{tri-zone net can be derived as one equation and try-on net can be concluded as other  from xb}
Training CrossVTON for tri-zone try-on relies on a large amount of quadruple data. The quadruple consists of: 1) a model picture $P_a$, where the model wears clothing of pattern a; 2) a model picture $P_b$, with the model wearing clothing of pattern $b$; 3) the clothing of pattern $b$; and 4) the tri-zone mask $M^{P_a \to P_b}$ that corresponds to substituting the clothing in $P_a$ with that in $P_b$. $P_a$ and $P_b$ share the same model, posture, and background, with the only variance being the clothing. Clothing a and clothing b can be either from the same category or different categories. A model trained in this fashion exhibits excellent generalization capabilities for both in-category and cross-category try-on.

However, existing virtual try-on datasets generally consist of pairs of model image and their corresponding clothing. This is merely a subset of the quadruple data, specifically the model picture $P_b$ and the clothing b. Even with the cooperation of the clothing model, it remains challenging to acquire $P_a$ and $M^{(P_a  \to P_b)}$. Therefore, the quadruple data can only be obtained through data construction methods. To align with the model training process, we denote the ground-truth model image during model training as $P_g$, the corresponding garment image as $G_g$, the model image constructed using $P_g$ and $G_c$ as $P_c$, and the ground-truth tri-zone mask for try-on the model image $P_c$ with $G_g$ as $M_{3g}^{P_{c} \to P_{g}}$. Consequently, the quadruple data is represented as $[P_{c}, P_{g}, G_{g}, M_{3g}^{Pc \to Pg}]$.

When training the first-stage model, $P_{c}$ is utilized as the input model image, and the garment image $G_{g}$ is employed as the input clothing. The model predicts the tri-zone mask $M_{3p}^{P_{c} \to P_{g}}$, with $M_{3g}^{P_{c} \to P_{g}}$ acting as the ground-truth for supervision during training. When training the second-stage model, the model takes the model image $P_{c}$, the ground-truth clothing image $G_{g}$, and the tri-zone mask $M_{3p}^{P_{c} \to P_{g}}$ predicted in first stage as inputs. Subsequently, the model predicts the try-on result $P_{r}$, with $P_{g}$ serving as the ground-truth for supervision in the training process.

\begin{table}[t!]
\centering
\resizebox{0.95\linewidth}{!}{
\begin{tabular}{ccccccccc}
\toprule
& & & \multicolumn{6}{c}{$P_{c}$} \\
\cmidrule(lr){4-9}
& & & \multicolumn{2}{c}{Upper} & \multicolumn{2}{c}{Dress} & \multicolumn{2}{c}{Lower} \\
\cmidrule(lr){4-9}
& & & Short & Long & Short & Long & Short & Long \\
\midrule
\multirow{6}{*}{$P_{g}$} & \multirow{2}{*}{Upper} & Short & \cellcolor{LightSalmon}IDM & \cellcolor{Moccasin}IDM\_S & \cellcolor{LightSalmon} & \cellcolor{LightSalmon} & \cellcolor{Gainsboro} & \cellcolor{Gainsboro} \\
& & Long & \cellcolor{Moccasin}IDM\_S & \cellcolor{LightSalmon}IDM & \multicolumn{2}{c}{\cellcolor{LightSalmon}\multirow{-2}{*}{IDM}} & \multicolumn{2}{c}{\cellcolor{Gainsboro}\multirow{-2}{*}{N/A}} \\
& \multirow{2}{*}{Dress} & Short & \cellcolor{LightCoral} & \cellcolor{LightCoral} & \cellcolor{LightSalmon}IDM & \cellcolor{Moccasin}IDM\_S & \cellcolor{LightCoral} & \cellcolor{LightCoral} \\
& & Long & \multicolumn{2}{c}{\cellcolor{LightCoral}\multirow{-2}{*}{CrossVTON}} & \cellcolor{Moccasin}IDM\_S & \cellcolor{LightSalmon}IDM & \multicolumn{2}{c}{\cellcolor{LightCoral}\multirow{-2}{*}{CrossVTON}} \\
& \multirow{2}{*}{Lower} & Short & \cellcolor{Gainsboro} & \cellcolor{Gainsboro} & \cellcolor{Moccasin} & \cellcolor{LightSalmon} & \cellcolor{LightSalmon} & \cellcolor{LightSalmon} \\
& & Long & \multicolumn{2}{c}{\cellcolor{Gainsboro}\multirow{-2}{*}{N/A}} & \cellcolor{Moccasin}\multirow{-2}{*}{IDM\_S} & \cellcolor{LightSalmon}\multirow{-2}{*}{IDM} & \multicolumn{2}{c}{\cellcolor{LightSalmon}\multirow{-2}{*}{IDM}} \\
\bottomrule
\end{tabular}
}
\vspace{-2mm}
\caption{Cross-Category data construction methods. Orange background denotes constructed by IDM. Yellow background denotes IDM improved by mask strategy. Red background denotes using our CrossVTON.}
\label{tab:Cross-Category data construction}
\vspace{-3mm}
\end{table}

\subsection{Progressive Learning Paradigm}
For the quadruple data used in training, the pattern differences between $P_{c}$ and $P_{g}$ should encompass as many scenarios as possible. As shown in Table 1, we categorize clothing based on types and lengths. This classification can generate diverse combinations of quadruples. Some of these quadruples can be constructed using existing mask-based methods or combined with some mask strategies. While for the remaining quadruples, it do not work. Nevertheless, these quadruples can be constructed by leveraging the try-on model trained with the previously constructed quadruples. Therefore, we propose a progressive learning paradigm (Fig.\ref{fig:main_framework}) that includes two rounds of data construction and two rounds of model training. First, we construct data using existing mask-based methods and their improved strategies. Then, we conduct the first-round model training using these data. Subsequently, we use the well-trained model for the second-round data construction. Finally, we perform the second-round training using all the constructed data.

\begin{figure}[t!]
    \centering
    % 第一张图片
    \includegraphics[width=0.40\textwidth]{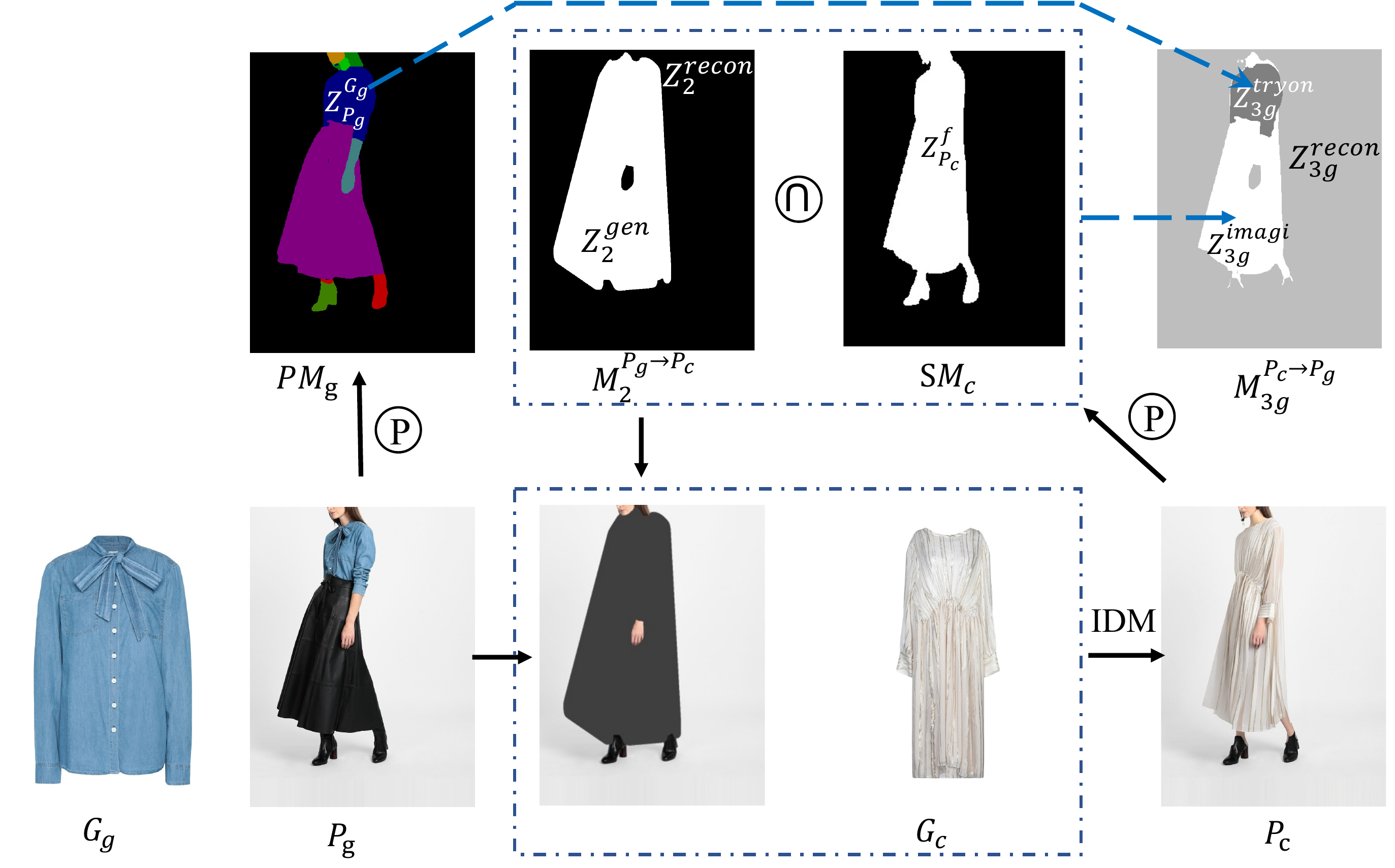}
    \vspace{-3mm}
    \caption{First round intra-category and any-to-dress data construction}
    \label{fig:construct-1}
    % 第二张图片
    \vspace{-3mm}
    \includegraphics[width=0.40\textwidth]{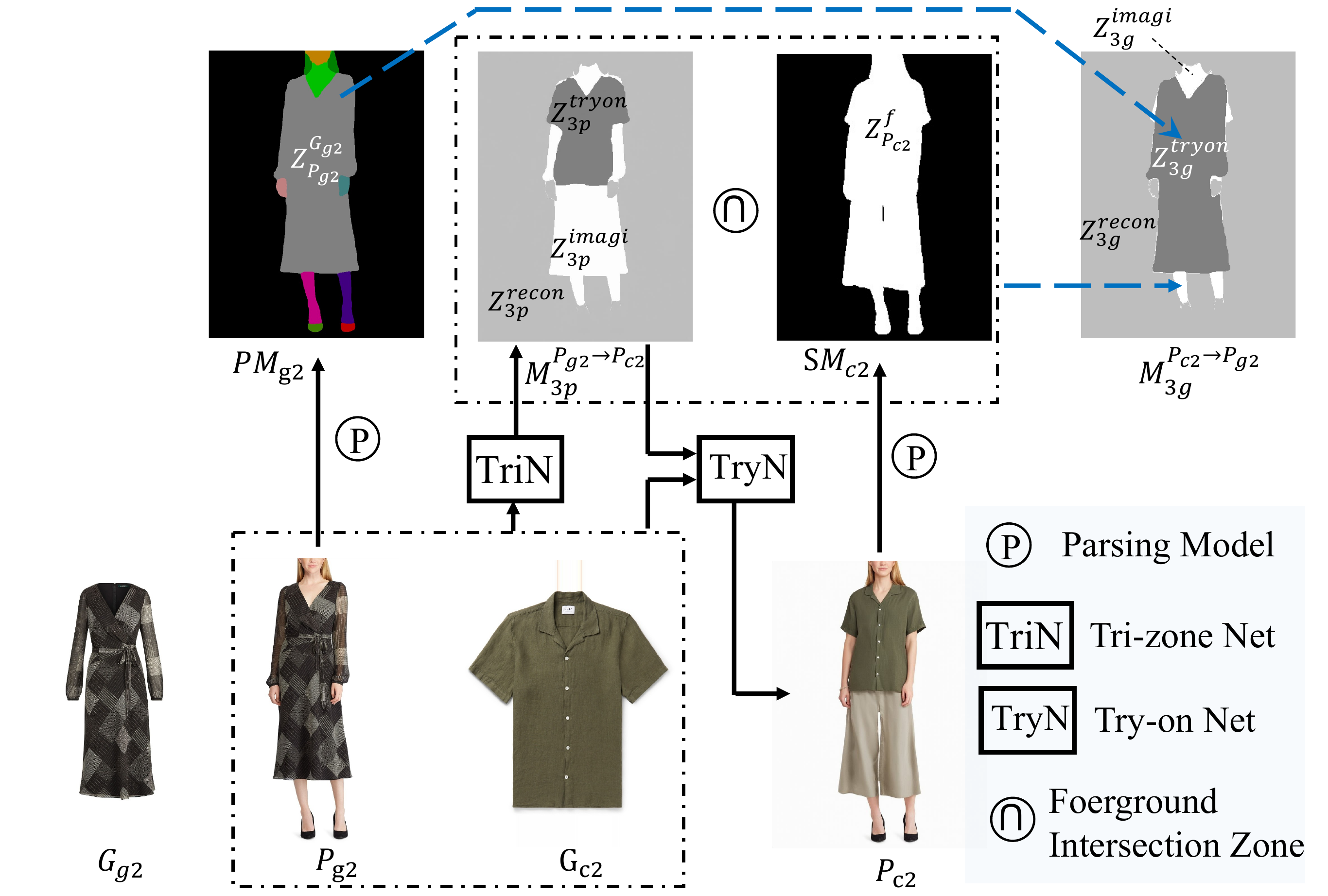}
    \vspace{-3mm}
    \caption{Second round dress-to-any data construction}
    \label{fig:construct-2}
    \vspace{-5mm}
\end{figure}

\subsubsection{First Round Data Construction}
\noindent\textbf{Intra-Category with match size and any-to-dress, by IDM.}
As shown by orange in Tab.\ref{tab:Cross-Category data construction}, when $G_{c}$ and $G_{g}$ fall into the same category and they have matched size (\textit{e.g.} both $G_{c}$ and $G_{g}$ are short T-shirt), or when $G_{c}$ can largely cover $G_{g}$ (\textit{e.g.} $G_{c}$ is dress and $G_{g}$ is short T-shirt), existing mask-based methods can already yield satisfactory results with specified mask. Therefore, we can directly utilize these methods to construct $P_{c}$ by inputting $P_{g}$ and $G_{c}$. In this paper, we employ IDM for data construction. The mask $M_{2}^{(P_{g} \to P_{c})}$ employed by IDM discriminates two types of regions. The reconstruction region $Z_{2}^{recon}$ denotes the area where $P_{c}$ remains consistent with $P_{g}$, while the generation region $Z_{2}^{gen}$ represents the area that is regenerated during the construction process. 

% \begin{figure*}[t!]
%     \centering
%     \includegraphics[width=0.75\textwidth]{Figure/rst_CCDC_CCGD.pdf}
%     \vspace{-3mm}
%     \caption{Visual results on CCDC and CCGD. Best viewed when zoomed in.}
%     \label{fig:rst_CCDC_CCGD}
%     % \vspace{-5mm}
% \end{figure*}

\noindent\textbf{Obtain tri-zone GT.}
As shown in Fig. \ref{fig:construct-1}, to construct the mask $M_{3g}^{(P_{c}\to P_{g})}$ required for training CrossVTON, we actually need to distinguish three types of regions. The try-on region $Z_{3g}^{tryon}$ represents the area of the clothing $G_{g}$ in $P_{g}$, which can be obtained by extracting the $G_{g}$ part $Z_{P_{g}}^{G_{g}}$ from the parsing map $PM_{g}$ of $P_{g}$, \textit{i.e.} $Z_{3g}^{tryon}=Z_{P_{g}}^{G_{g}}$. The imagination region $Z_{3g}^{imagi}$ represents the area that the model needs to imagine and supplement reasonably. It can be obtained by taking the intersection of the generation region $Z_{2}^{gen}$ during construction and the foreground area $Z_{P_c}^{f}$ of $P_{c}$ segment mask $SM_c$, and then subtracting the intersection with the try-on region $Z_{3g}^{tryon}$, \textit{i.e.} $Z_{3g}^{imagi}=(Z_{2}^{gen} \cap Z_{P_c}^{f})-Z_{3g}^{tryon}$. The reconstruction region $Z_{3g}^{recon}$ represents the area that should be consistent with the input $P_{c}$, it is the area excluding $Z_{3g}^{tryon}$ and $Z_{3g}^{imagi}$. 

\noindent\textbf{Intra-category with mismatch size, by IDM\_S.}
As indicated by red in Tab.\ref{tab:Cross-Category data construction}, these size-mismatch cases can't be handle by IDM. Therefore, we design dedicated data construction for such scenarios. The following uses tops as an example for illustration. For the scenario where $G_{g}$ is a shorter top and $G_{c}$ is a longer top, we leverage the densepose of model $P_{g}$ to shift the lower boundary of $Z_{2}^{gen}$ downward, thereby obtaining $Z_{2}^{gen \downarrow}$ . Based on the new mask $M_{2}^{(P_{gt}\to P_{ct}) \downarrow}$ , we employ IDM to conduct the virtual try-on process, resulting in the model image $P_{c}^{\downarrow}$ depicting the model wearing a longer top. For the case where $G_{g}$ is a long top and $G_{c}$ is a short top, we randomly shift the lower boundary of $Z_{2}^{gen}$ upward to get $Z_{2}^{gen \uparrow}$. Subsequently, based on the new mask $M_{2}^{(P_{g}\to P_{c}) \uparrow}$ , we utilize IDM to conduct the try-on process, thus obtaining the model $P_{ct}^{\uparrow'}$  wearing a shorter top. Nevertheless, the content within the corresponding area of $Z_{2}^{gen} - Z_{2}^{gen \uparrow}$ in $P_{c}^{\uparrow'}$  remains the same as that in $P_{g}$. Ultimately, we employ an inpainting model to complete this area, resulting in a reasonable $P_{c}^{\uparrow}$. The construction of $M_{3g}^{(P_{c}\to P_{g})}$ follows the same approach as the IDM-based construction method.

\begin{table*}[htb] 
    \centering
    % \scriptsize
    \resizebox{\textwidth}{!}{%
    \begin{tabular}{lcccccccccccc}
        \toprule 
        \multirow{2}{*}{Methods} & \multicolumn{6}{c}{DressCode} & \multicolumn{6}{c}{VITON-HD} \\
        \cmidrule(lr){2-7} \cmidrule(lr){8-13}
        & \multicolumn{4}{c}{Paired} & \multicolumn{2}{c}{Unpaired} & \multicolumn{4}{c}{Paired} & \multicolumn{2}{c}{Unpaired} \\
        \cmidrule(lr){2-5} \cmidrule(lr){6-7} \cmidrule(lr){8-11} \cmidrule(lr){12-13}
        & SSIM $\uparrow$ & LPIPS $\downarrow$ & FID $\downarrow$ & KID $\downarrow$  & FID $\downarrow$ & KID $\downarrow$ & SSIM $\uparrow$ & LPIPS $\downarrow$  & FID $\downarrow$ & KID $\downarrow$ & FID $\downarrow$ & KID $\downarrow$ \\
        \midrule
        LaDI-VTON (2023) & 0.9025 &	0.0719&	4.8636&	1.5580&	6.8421&	2.3345	&	0.8763&	0.0911&	6.6044&	1.0672	&9.4095& 1.6866 \\
        StableVTON (2024) &-	&-&	-	&-	&-	&-		&0.8665	&0.0835	&6.8581&	1.2553&	9.5868	&1.4508 \\

        OOTDiffusion (2024) & 0.8975&	0.0725	&3.9497	&0.7198&	6.7019&	1.8630	&	0.8513&	0.0964&	6.5186&	{0.8961}	& 9.6733	&1.2061\\
        IDM-VTON (2024) &\textbf{0.9228}&	0.0478	&3.8001	&1.2012	&5.6159	&1.5536		&\textbf{0.8806}	&0.0789	&6.3381	& 1.3224	&9.6114	& 1.6387\\
        CatVTON (2024) & 0.9011 &	0.0705 & 3.2755 &{0.6696} &5.4220 &1.5490   &
        % & {0.8922} & {0.0485} & {3.992} & {0.8180}  & {6.137} & {1.403} 
        {0.8694} & {0.0970} & {6.1394} & {0.9639} 
        & {9.1434} & {1.2666}\\
        \midrule
        {CrossVTON (Ours)} & {0.9130} & \textbf{0.0470} & \textbf{2.7515} & \textbf{0.5229} & \textbf{5.0465} & \textbf{1.2368} & 
        {0.8550} & \textbf{0.0786} & \textbf{5.7171} & \textbf{0.8804} & \textbf{8.6596} & \textbf{0.8120} \\

        \bottomrule
    \end{tabular}
    }
   \vspace{-3mm}
    \caption{\small Quantitative results on DressCode and VITON-HD datasets.
    % Bold denotes the best score for each metric.
    }
    \label{tab:results-convention}
    \vspace{-4mm}
\end{table*}

\begin{figure*}[t!]
    \centering
    \includegraphics[width=0.70\textwidth]{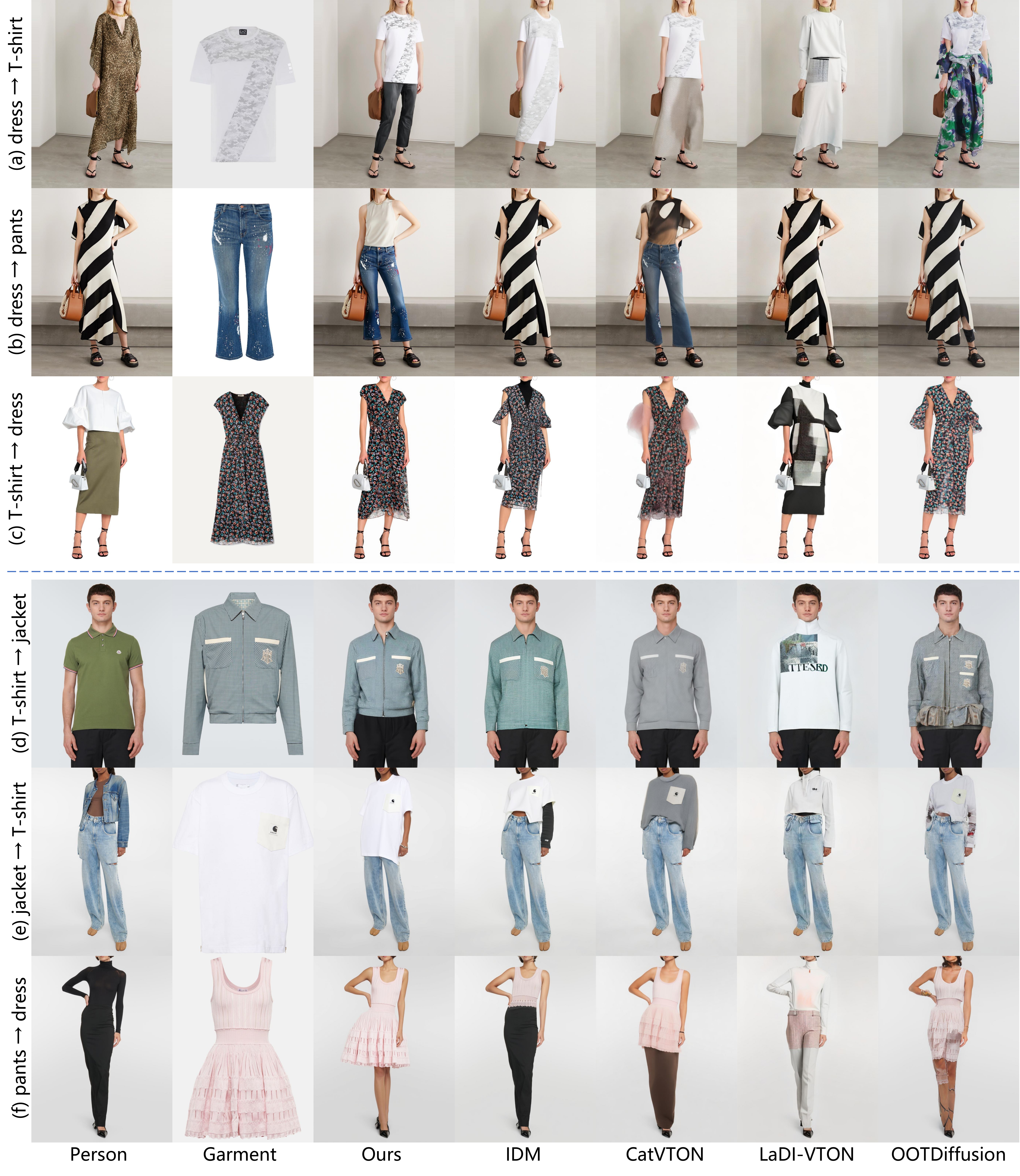}
    \vspace{-4mm}
    \caption{Visual results on CCDC (top) and CCGD (bottom). Best viewed when zoomed in.}
    \label{fig:rst_CCDC_CCGD}
    \vspace{-5mm}
\end{figure*}

\subsubsection{First Round Model Training}
As shown by red in Tab.\ref{tab:Cross-Category data construction}, for these categories, existing methods are unable to construct an upper-garment or lower-garment model from a dress-wearing model. Specifically, the constructed model image is not reasoning or it is exactly the original model image. However, we successfully performed the reverse data construction, namely constructing a dress-wearing model from an upper-garment or lower-garment model. Therefore, we can utilize these quadruple data to train our CrossVTON. Once trained, CrossVTON can generate reasonable try-on results when provided with a dress model and an upper-garment or lower-garment.

\subsubsection{Second Round Data Construction}
\noindent\textbf{Dress-to-any, By CrossVTON.} 
To distinguish from the first round, elements used in the second round in the following text are marked with subscript \textit{2}. As shown in Fig.\ref{fig:construct-2}, based on the CrossVTON obtained from the first round training, using the dress-wearing model $P_{g2}$ and the upper-garment/lower-garment $G_{c2}$ as inputs, we can batch-construct $P_{c2}$ with the upper-garment/lower-garment on. The mask $M_{3p}^{(P_{g2}\to P_{c2})}$ predicted by the first stage of CrossVTON consists of three types of regions: the try-on region $Z_{3p} ^{tryon}$, the reconstruction region $Z_{3p}^{recon}$, and the imagination region $Z_{3p}^{imagi}$. When constructing $M_{3g}^{(P_{c2}\to P_{g2})}$, we take the parsed $G_{c2}$ region $Z_{P_{g2}}^{G_{g2}}$ which from the parsing map $PM_{g2}$ as $Z_{3g}^{tryon}$, \textit{i.e.} $Z_{3g}^{tryon}=Z_{P_{g2}}^{G_{g2}}$. The imagination area $Z_{3p}^{imagi}$ can be obtained using $Z_{3p}^{tryon}$, $Z_{3p}^{imagi}$ and $Z_{P_c}^{f}$, \textit{i.e.} $Z_{3g}^{imagi}=(Z_{3p}^{tryon} \cup Z_{3p}^{imagi} \cap Z_{P_c}^{f})-Z_{3g}^{tryon}$. While the reconstruction region $Z_{3p}^{recon}$ is the remaining area.

\subsubsection{Second Round Model Training}
By training the model using all the data generated in the previous two stages, we can acquire the final model. This final model is able to support cross-category and intra-category (wether size match or not) try-on task.

\section{Experiments}

\noindent\textbf{Dataset.} 
We trained our model on two widely adopted virtual try-on datasets: VITON-HD \cite{choi2021viton} and DressCode \cite{morelli2022dress}. By combining model image and garment image from different category of DressCode, we construct a Cross-Category DressCode test set, namely CCDC which include 7200 test pairs.
% The VITON-HD dataset consists of 13,679 upper-body image pairs, split into 11,647 training pairs and 2,032 testing pairs. The DressCode dataset includes 48,392 training pairs and 5,400 testing pairs, featuring full-body portraits and three garment categories: upper-body, lower-body, and dresses. Both datasets are provided at a resolution of 1024×768.
To more effectively assess the performance of cross-category clothing transfer, a dedicated test set for this purpose, named the Cross-Category Garment Dataset (CCGD), has been meticulously assembled. The CCGD encompasses eight distinct categories: long-tops, short-tops, long dresses, short dresses, long pants, short pants, long skirts, and short skirts. In total, it comprises 400 model-garment pairs, 50 pairs for each category. The model and garment items from different categories are combined to compose 2000 test set pairs.
% zicai dataset?
% The DressCode dataset provides garment images alongside corresponding ground truth human images, offering a robust foundation for training and evaluation in cross-category virtual try-on scenarios.

\noindent\textbf{Baselines.} 
% To evaluate our approach,
We compare CrossVTON with five SOTA
% state-of-the-art 
diffusion-based virtual try-on methods: LaDI-VTON \cite{morelli2023ladi}, StableVTON \cite{kim2024stableviton}, OOTDiffusion \cite{xu2024ootdiffusion}, IDM-VTON \cite{choi2024improving}, and CatVTON \cite{zeng2024cat}. 
While these methods excel in standard virtual try-on tasks, they fail to address cross-category scenarios with significant size mismatches.

\noindent\textbf{Evaluation.} 
% For model and garment within the same garment category, 
Similar to other methods, we use SSIM 
% (Structural Similarity Index Measure) 
and LPIPS
% (Learned Perceptual Image Patch Similarity)
% to measure the reconstruction quality
, and FID 
% (Frechet Inception Distance)
and KID 
% (Kernel Inception Distance)
to evaluate the authenticity of the generated images. For the pair setting with ground truth (GT), we report the above four indicators. For the unpair setting without GT, we report FID and KID.
% For the cross-category test sets CCDC and CCGD, metrics that can reflect the accuracy of the try-on results are essential. 
Through extensive experimentation, we discovered that given a model, clothing, and the outcome of a try-on model, Qwen-VL-Max can precisely determine whether the clothing transfer result is correct. Consequently, we utilize the accuracy rate provided by this model (\textit{i.e.,} the ratio of the number of correct results to the total number of samples in the entire test set) to evaluate the model's proficiency in cross-category try-on. 
% Additionally, FID and KID are employed to assess the plausibility of clothing transfer.

\noindent\textbf{Implementation details.} 
We utilize the official training set of VITON-HD and DressCode to construct training quadruple VITONHD-CT and DressCode-CT respectively. The former only has intra-category quadruple while the later has intra- and cross-category quadruple. To validate the performance of cross-category clothing transfer, we combine VTONHD-CT and DressCode-CT for training. Then, we conduct evaluations on the cross-category test sets CCDC and CCGD. The qualitative results are also based on this model. All models are trained uniformly with a batch size of 32, employing the AdamW optimizer at a learning rate of $ 3\times 10^{-5}$.

% Since VITON-HD solely concentrates on the upper body, it is only capable of constructing data within a single category. We utilize the official training set to create the training quadruple set named VTONHD-ct-train. DressCode encompasses three categories and can be employed to generate cross-category data. Based on the official training set, we construct the cross-category training set named DressCode-ct-train. Additionally, we construct the cross-category test dataset named DressCode-ct-test using the official test set. To validate the performance of cross-category clothing transfer, we combine VTONHD-ct-train and DressCode-ct-train for training. Then, we conduct evaluations on the cross-category test sets DressCode-ct-test and CCGD. The qualitative results are also based on this model. All models are trained uniformly with a batch size of 32, employing the AdamW optimizer at a learning rate 397
% of 3e-5.

% \begin{figure*}[htb]
%     \centering
%     \includegraphics[width=0.85\textwidth]{Figure/rst_CCDC_CCGD.pdf}
%     \vspace{-3mm}
%     \caption{Visual results on CCDC and CCGD. Best viewed when zoomed in.}
%     \label{fig:rst_CCDC_CCGD}
%     \vspace{-5mm}
% \end{figure*}

\subsection{Quantitative results}
As depicted in Tab.~\ref{tab:results-convention}, for try-on task within the same category, in comparison with several recent methods, our approach significantly outperforms all baselines. This verifies that our model, which is primarily engineered for cross-category try-on, also attains the SOTA level in conventional try-on tasks. For cross-category try-on, as illustrated in Tab.~\ref{tab:rst_CCDC_CCGD} CrossVTON has an advantage of 14\% and 16\% respectively over the second-best method On CCDC and CCGD, demonstrating its absolute leadership in the cross-category clothing transfer task. 

% \begin{table}[t!]
% \centering
% % \scriptsize
% \footnotesize
% \begin{tabular}{lccc}
% \hline
% Methods & FID $\downarrow$ & KID $\downarrow$ & Acc $\uparrow$ \\
% \hline
% LaDI-VTON & 34.53 & 23.11 & 12.54 \\
% IDM-VTON & 18.51 & 9.31 & 45.04 \\
% OOTDiffusion & 28.69 & 16.52 & 40.72 \\
% CatVTON & 23.01 & 14.22 & 55.23 \\
% Ours & \textbf{9.35} & \textbf{3.11} & \textbf{69.11} \\
% \hline
% \end{tabular}
% \vspace{-3mm}
% \caption{Quantitative results on Cross-Category DressCode.}
% \label{tab:results-CCDC}
%   \vspace{-3mm}
% \end{table}

% \begin{table}[t!]
% \centering
% \footnotesize
% \begin{tabular}{lccc}
% \hline
% Methods & FID $\downarrow$ & KID $\downarrow$ & Acc $\uparrow$ \\
% \hline
% LaDI-VTON & 52.2 & 20.16 & 12.65 \\
% IDM-VTON & 39.99 & 8.22 & 45.85 \\
% OOTDiffusion & 52.72 & 16.7 & 33.05 \\
% CatVTON & 46.86 & 14.02 & 41.75 \\
% Ours & \textbf{33.07} & \textbf{3.46} & \textbf{61.75} \\
% \hline
% \end{tabular}
% \vspace{-3mm}
% \caption{Quantitative results on CCGD.}
% \label{tab:results-CCGD}
%   \vspace{-3mm}
% \end{table}

%%%%
\begin{table}[t!] 
    \centering
    \resizebox{0.5\textwidth}{!}{%
    \begin{tabular}{lcccccc}
        \toprule 
        \multirow{2}{*}{Methods}  & \multicolumn{3}{c}{ Cross-Category DressCode} & \multicolumn{3}{c}{CCGD} \\
        \cmidrule(lr){2-4} \cmidrule(lr){5-7}
         & FID $\downarrow$ & KID $\downarrow$ & Acc $\uparrow$ & FID $\downarrow$ & KID $\downarrow$  & Acc $\uparrow$ \\
        \midrule
        LaDI-VTON & 34.53 & 23.11& 12.54  & 52.2 &20.16 &12.65    \\
        IDM-VTON & 18.51&  9.31 & 45.04 & 39.99 & 8.22 & 45.85   \\
        OOTDiffusion & 28.69 &16.52 &40.72 & 52.72& 16.7 &33.05  \\
        CatVTON &  23.01& 14.22& 55.23&  46.86& 14.02& 41.75 \\
        \midrule
        {Ours} & \textbf{9.35} & \textbf{3.11} & \textbf{69.11} & \textbf{33.07} & \textbf{3.46} & \textbf{61.75}  \\
        \bottomrule
    \end{tabular}
    }
    \vspace{-3mm}    \caption{\small Quantitative results on Cross-Category DressCode and CCGD dataset.}
    \label{tab:rst_CCDC_CCGD}
    \vspace{-3.5mm}
\end{table}

%%%%

\begin{table}[t!]
\centering
\footnotesize
\begin{tabular}{lccc}
\hline
Methods & FID $\downarrow$ & KID $\downarrow$ & Acc $\uparrow$ \\
\hline
Ours$_{\textrm{w/o Tri-zone}}$ & 36.29 & 4.99 & 57.05 \\
Ours$_{\textrm{merge}}$ & 33.37 & 4.06 & 58.40 \\
Ours & \textbf{33.07} & \textbf{3.46} & \textbf{61.75} \\
\hline
\end{tabular}
\vspace{-3mm}
\caption{Ablation study on CCGD.}
\label{tab:results-ablation}
  \vspace{-4mm}
\end{table}

\subsection{Qualitative results}
Fig.\ref{fig:rst_CCDC_CCGD} showcases a comparison of the visual effects of CrossVTON and other methods on CCDC and CCGD. Evidently, our method exhibits remarkable advantages across diverse cross-category scenarios. (a) and (b) illustrate that CrossVTON is capable of managing try-on tasks like transforming a skirt into an upper-garment or lower-garment, which needs sound imagination capabilities. In contrast, other methods are incapable of handling these tasks. (c), and (f) reveal that CrossVTON can effectively preserve the pattern of the dress, regardless of the clothing initially worn by the model. In contrast, the results of other methods are influenced by the original clothing on the model, with the patterns differing substantially from the provided clothing. (d) and (e) demonstrate that when the category and length of the model's upper-garment vary from those of the given upper-garment, CrossVTON can consistently match the provided clothing adeptly, while other methods fall short in appropriately dealing with such cases.

\begin{figure}[t!]
    \centering
    \includegraphics[width=0.40\textwidth]{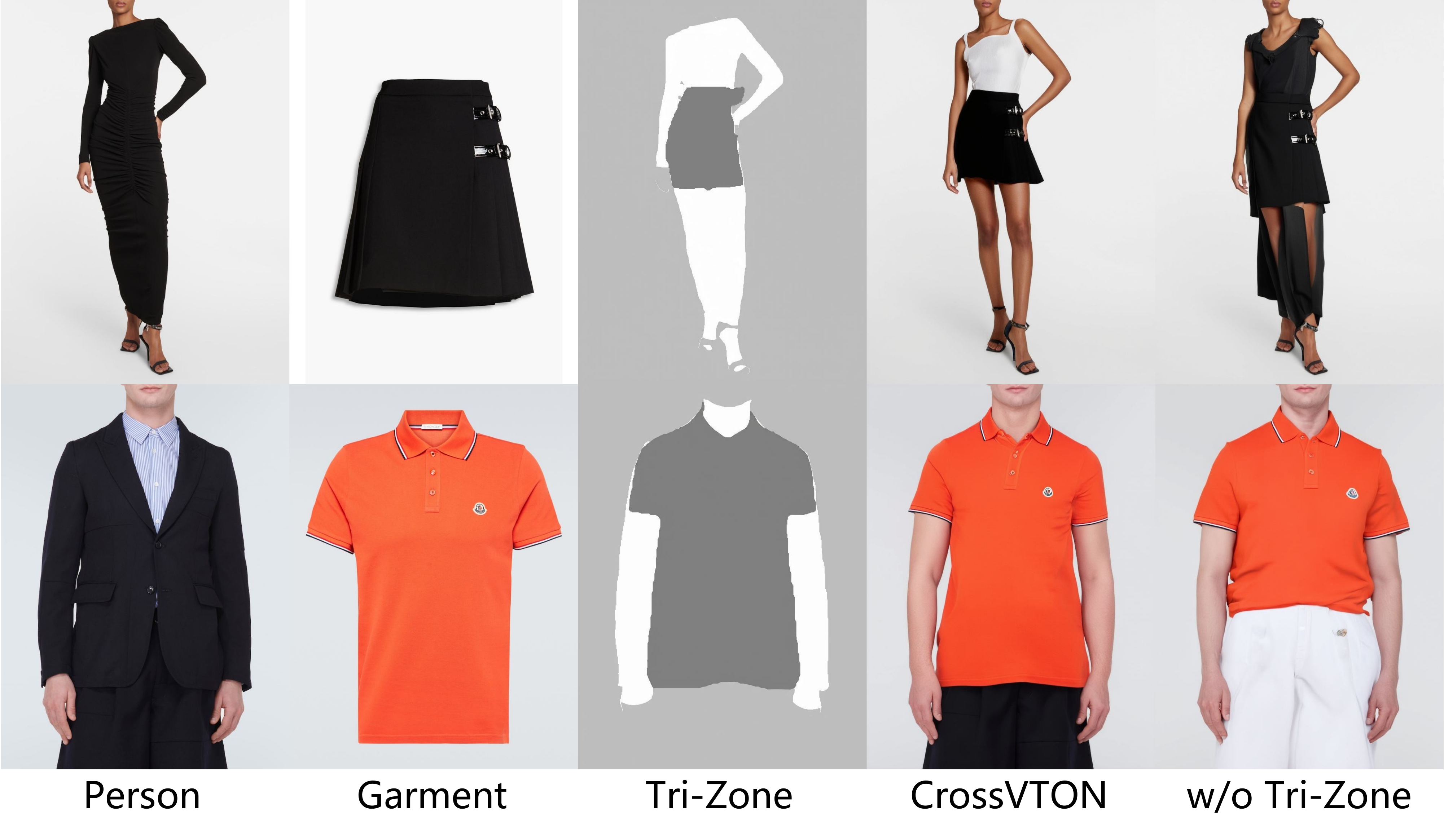}
    \vspace{-4mm}
    \caption{Ablation study on w/wo tri-zone prior}
    \label{fig:ab1}
    \vspace{-4mm}
\end{figure}

\begin{figure}[t!]
    \centering
    \includegraphics[width=0.40\textwidth]{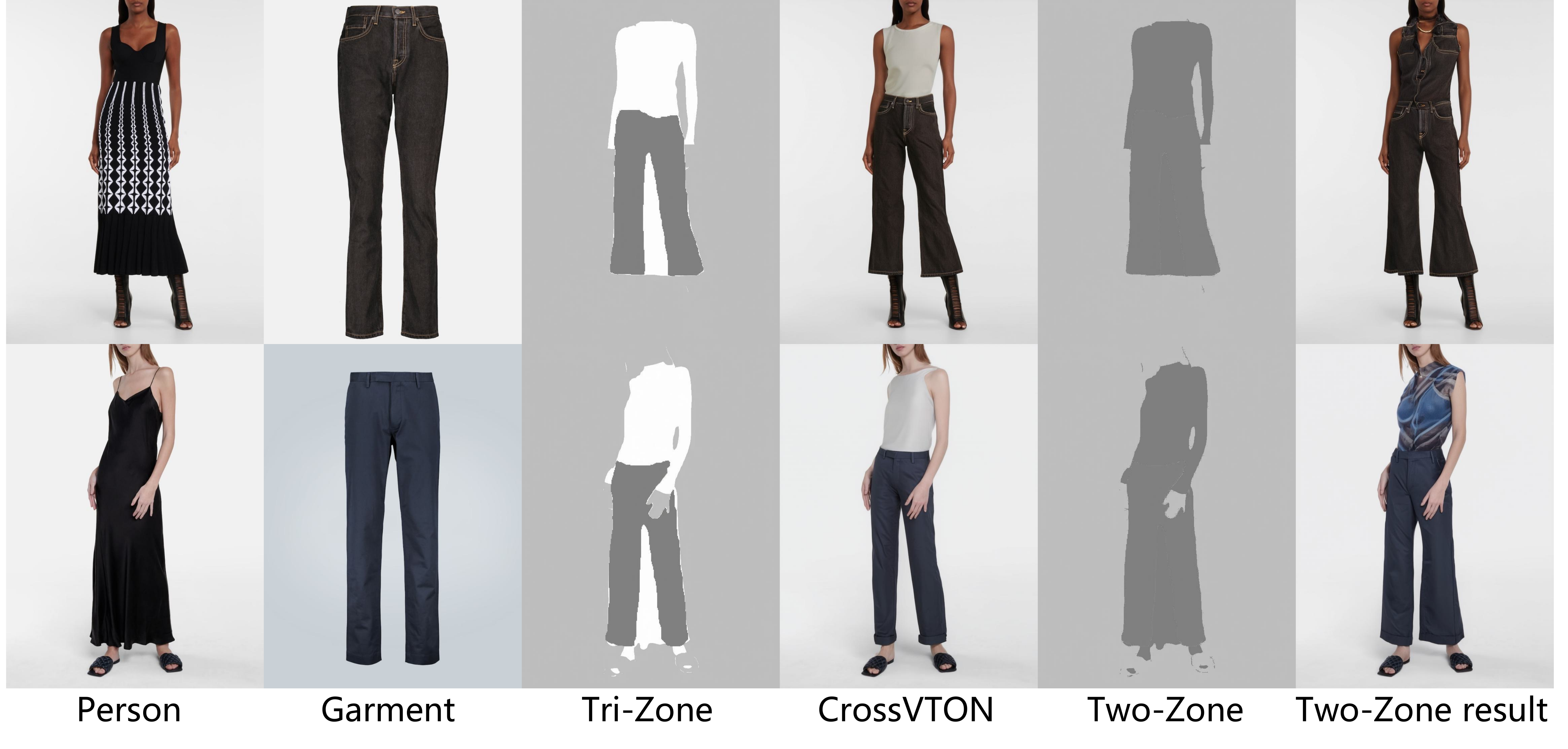}
    \vspace{-4mm}
    \caption{Ablation study on tri-zone and binary prior.}
    \label{fig:ab2}
    \vspace{-5mm}
\end{figure}

\subsection{Ablation Study}

To validate the role of the three-region prior, we remove the first stage of CrossVTON and directly utlizing $[P_{ct}, P_{gt}, C_{gt}]$ to train a mask-free Try-on Net. As presented in the tab.\ref{tab:results-ablation}, compared with CrossVTON, the accuracy Acc drops significantly by 4.7\%, and both FID and KID also declines substantially. Nevertheless, our mask-free model still holds a remarkable 11.2\% edge over IDM. This clearly demonstrates the effectiveness of our progressive learning paradigm. As depicted in the fig. \ref{fig:ab1}, the tri-zone prior can enhance the stability of the model.
To validate the difference between the tri-zone and binary prior, we merge the try-on region and the imagination region of Tri-zone mask to form a binary prior and directly used for Try-on Net. As shown in the tab. \ref{tab:results-ablation}, the binary prior leads to decrease on accuracy compared to the tri-zone prior, but still has a significant advantage over other methods. This indicates the tri-zone prior can provide more information to help try-on. It also demonstrates robustness of Try-on Net for different mask. The visual comparison in Fig. \ref{fig:ab2} shows the impact of the different priors on the results.

% \begin{figure*}[htb]
%     \centering
%     \includegraphics[width=0.89\textwidth]{Figure/rst_CCDC.pdf}
%     \vspace{-3mm}
%     \caption{Visual results on CCDC. Best viewed when zoomed in.}
%     \label{fig:rst_CCDC}
%     \vspace{-5mm}
% \end{figure*}

% \begin{figure*}[htb]
%     \centering
%     \includegraphics[width=0.89\textwidth]{Figure/rst_CCGD.pdf}
%     \vspace{-3mm}
%     \caption{Visual results on CCGD. Best viewed when zoomed in.}
%     \label{fig:rst_CCGD}
%     \vspace{-5mm}
% \end{figure*}

\section{Conclusion}
% We propose tri-zone priors to mimic the logic reasoning to distinguish different functionalities of various zones (\textit{i.e.,} try-on, reconstruction, or imagination area) after considering the cross-category inputs. Specifically, to embed the model with the reasoning ability on the cross-category cases, we build up an iterative data constructor to cover main occasions including the intra-category, any-to-dress and dress-to-any virtual try-on. Such an iterative data constructor, our CrossVTON is progressively trained guided by tri-zone priors to get the power for cross-category virtual try-on.

We propose novel tri-zone priors to emulate logical reasoning in distinguishing the distinct functionalities of various zones (\textit{i.e.,} try-on, reconstruction, or imagination zones) when considering cross-category inputs. Specifically, to endow the model with reasoning capabilities for cross-category scenarios, we also develop an iterative data constructor that encompasses key cases, including intra-category, any-to-dress, and dress-to-any virtual try-on. Through this iterative data construction process, our CrossVTON is progressively trained under the guidance of tri-zone priors, thereby acquiring the capability for cross-category virtual try-on.

\bibliographystyle{named}
\bibliography{ijcai25}

\end{document}

\maketitle

\section{Overview}
\label{sec:Overview}
In this supplementary document, we present additional details and results to complement the paper. 

\subsection{Evaluation Detail of Acc-Qwen}

\begin{figure}[htb]
    \centering
    \includegraphics[width=0.88\linewidth]{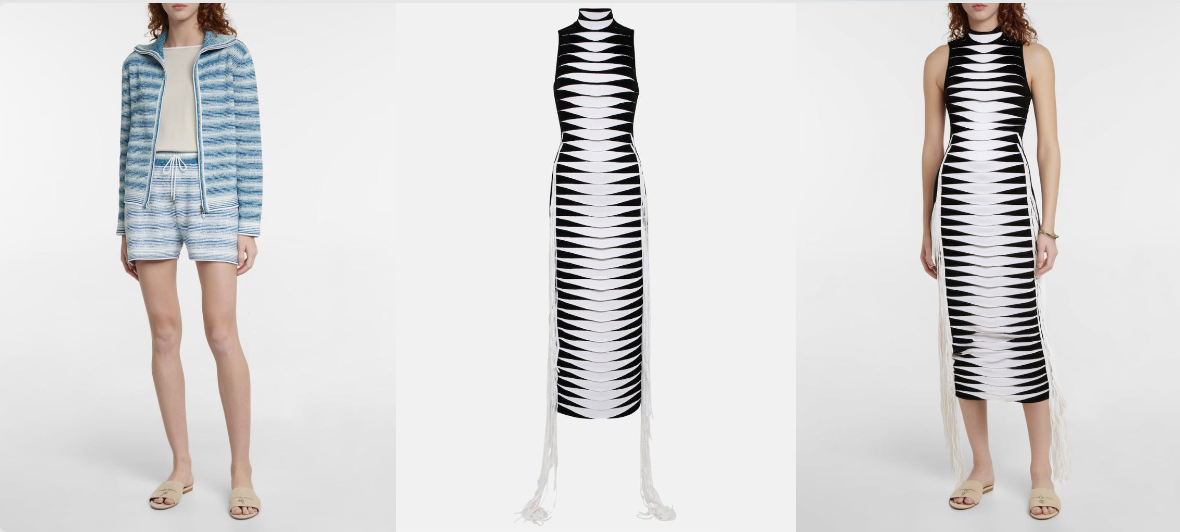}
    \vspace{-2mm}
    \caption{Input image for Qwen-VL-Max.}
    \vspace{-4mm}
    \label{fig:qwen_input}
\end{figure}

As shown in Fig.\ref{fig:qwen_input}, the input model image, input garment image and try-on result are spliced together as input of Qwen-VL-Max. The Prompt for the Multimodal Large Language Model as follows:

\textit{I used the virtual try-on algorithm to replace the model‘s garment in the left-hand image with the garment in the middle. Then produced the output on the right. If the overall model image on the right is reasonable and matches the type and style of the middle-image clothing, it’s considered reasonable. If the output image is the same as the input model image or the garment of output is not consistent with the middle image, the output is unreasonable. You only need to judge if it's reasonable. Reply "reasonable" if it is, and "unreasonable" if not.} 

Then the model will give its judgment "reasonable" or "unreasonable".

\subsection{More Qualitative Results on CCDC and CCGD}

\begin{figure*}[t]
    \centering
    \includegraphics[width=0.90\textwidth]{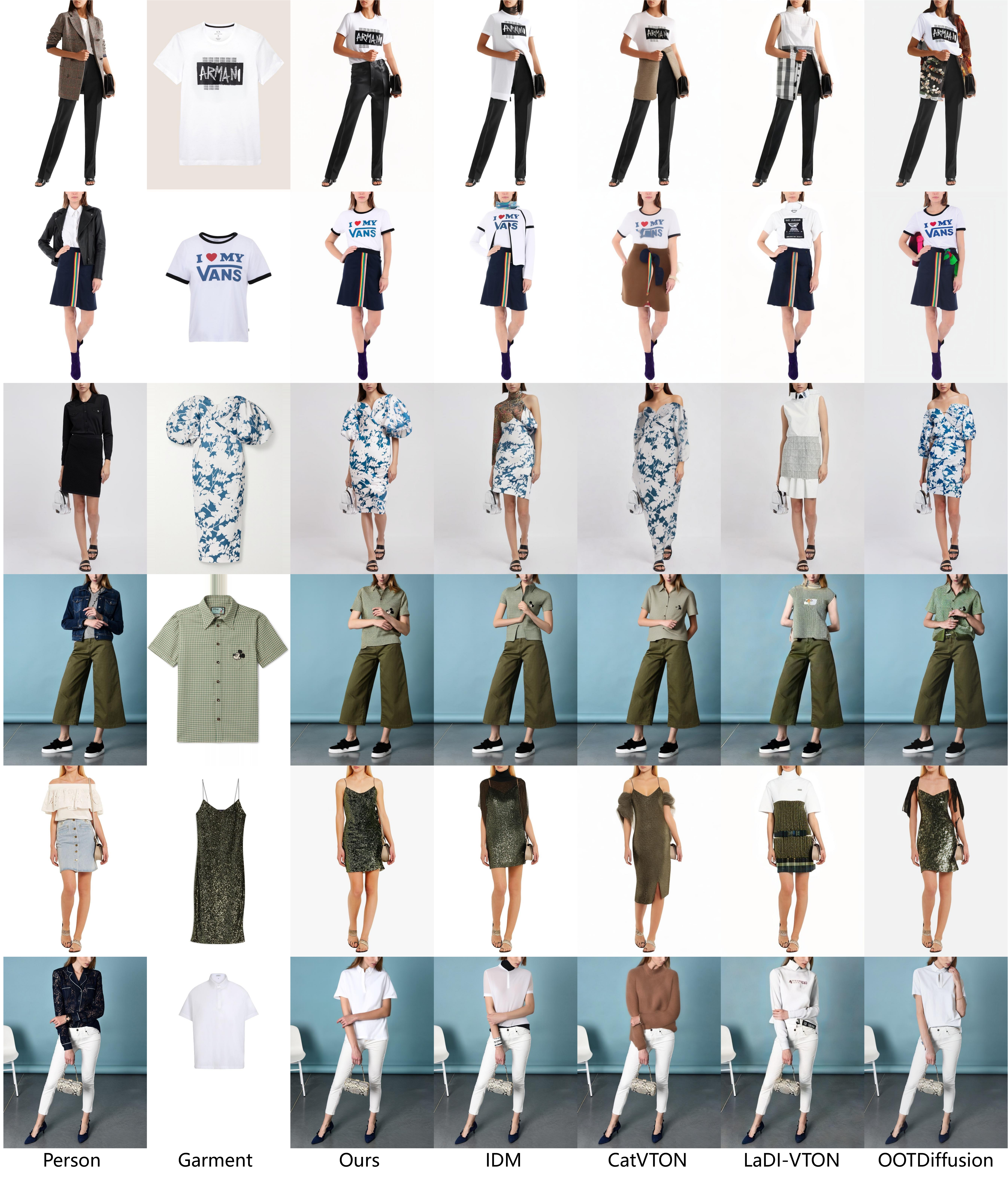}
    % \vspace{-4mm}
    \caption{More visual results on CCDC. Best viewed when zoomed in.}
    \label{fig:rst_CCDC_1}
    % \vspace{-5mm}
\end{figure*}

\begin{figure*}[t]
    \centering
    \includegraphics[width=0.90\textwidth]{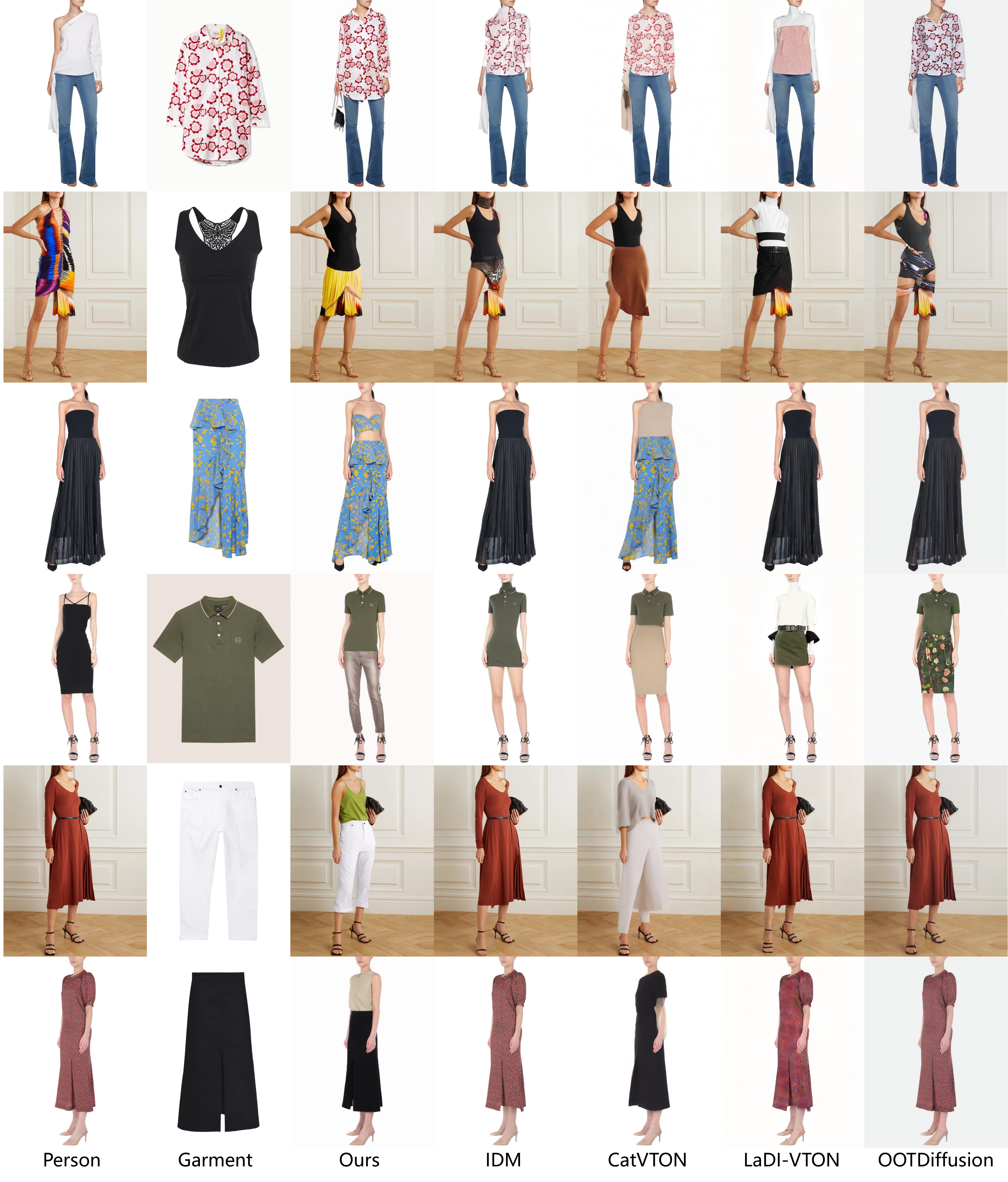}
    % \vspace{-4mm}
    \caption{More visual results on CCDC. Best viewed when zoomed in.}
    \label{fig:rst_CCDC_2}
    % \vspace{-5mm}
\end{figure*}

\begin{figure*}[t]
    \centering
    \includegraphics[width=0.90\textwidth]{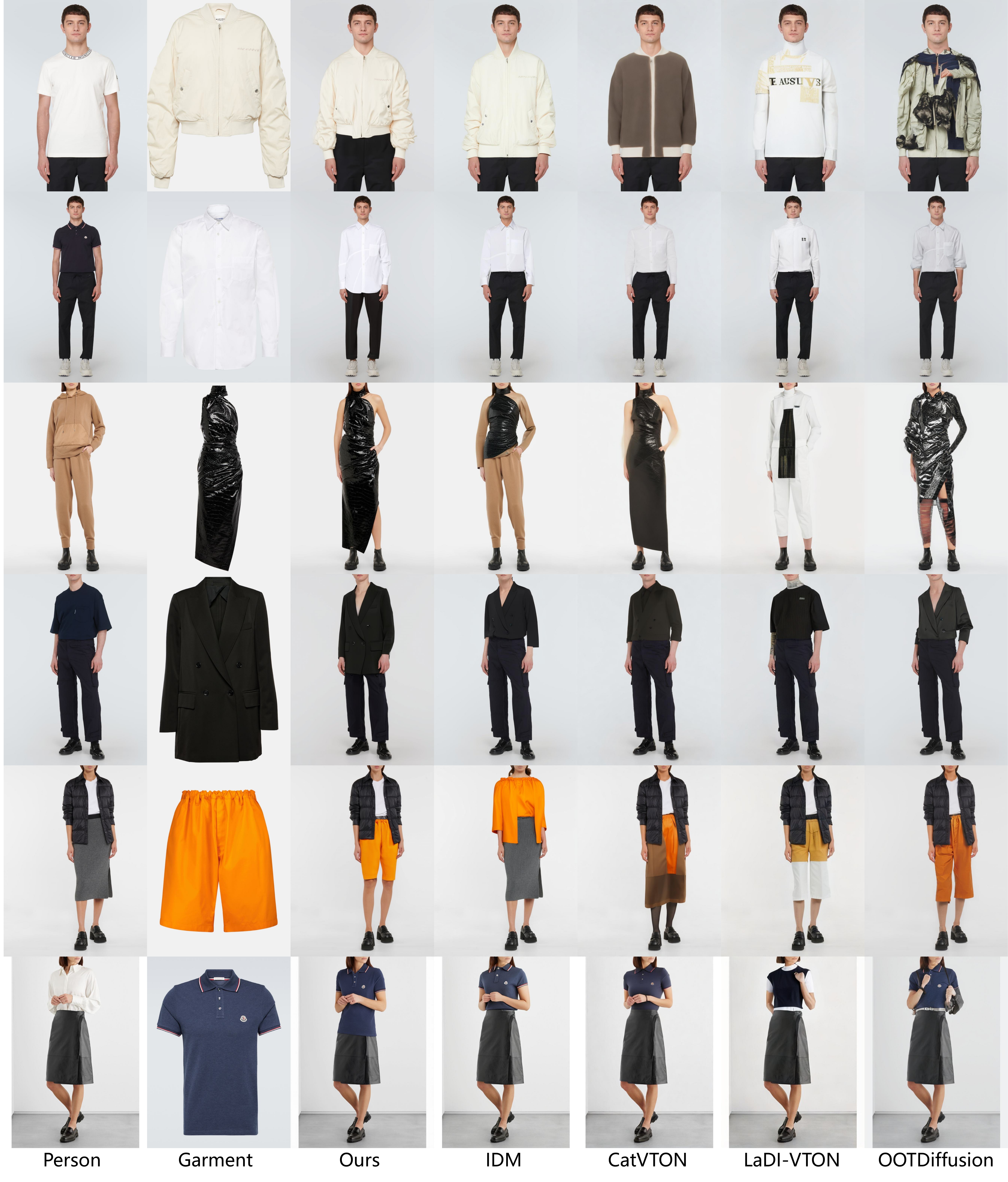}
    % \vspace{-4mm}
    \caption{More visual results on CCGD. Best viewed when zoomed in.}
    \label{fig:rst_CCGD_1}
    % \vspace{-5mm}
\end{figure*}

\begin{figure*}[t]
    \centering
    \includegraphics[width=0.90\textwidth]{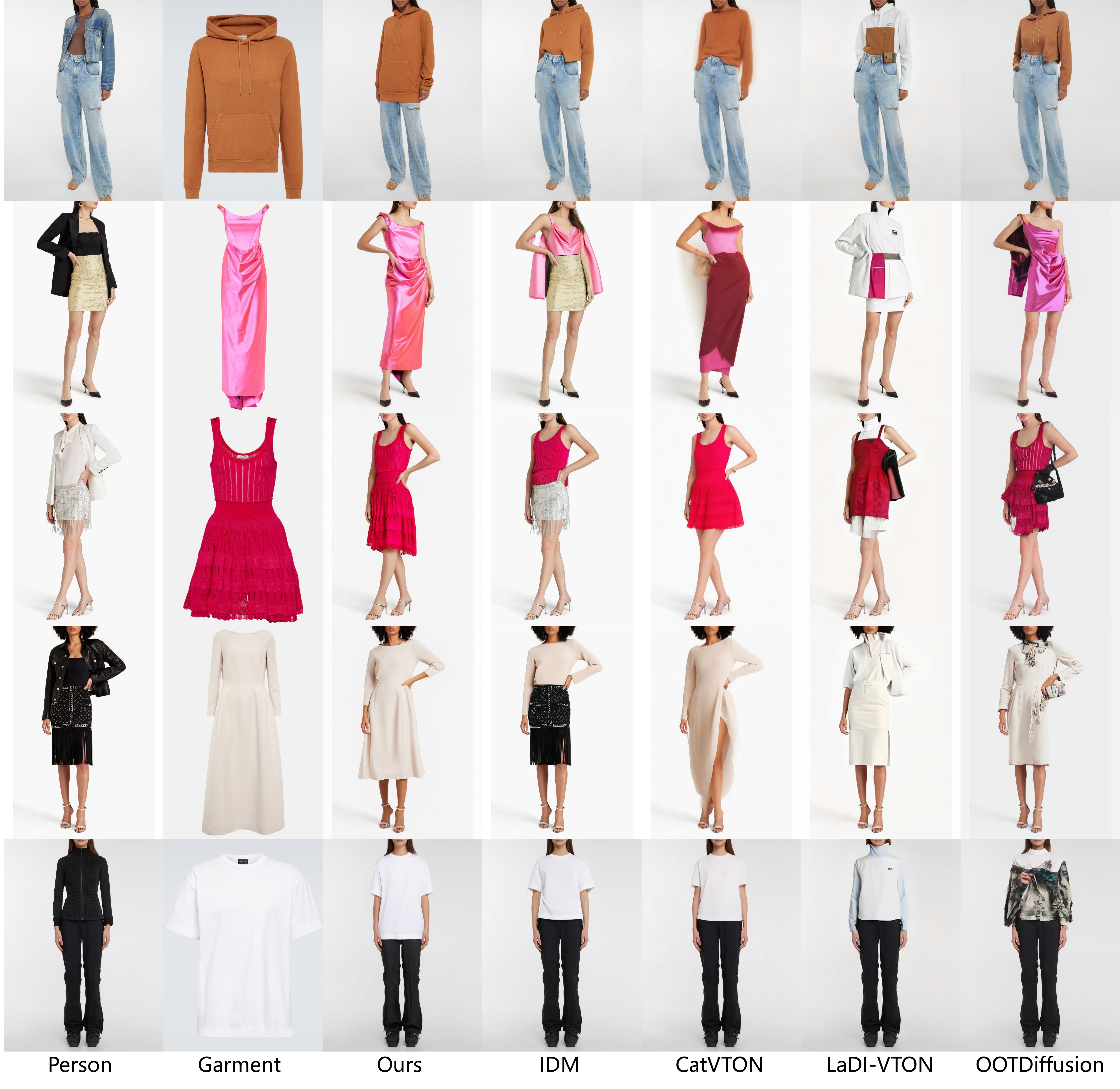}
    % \vspace{-4mm}
    \caption{More visual results on CCGD. Best viewed when zoomed in.}
    \label{fig:rst_CCGD_2}
    % \vspace{-5mm}
\end{figure*}